%% file: main.tex
\documentclass[10pt,twocolumn,letterpaper]{article}

\usepackage{cvpr}              %

\input{preamble}
\definecolor{cvprblue}{rgb}{0.21,0.49,0.74}
\usepackage[pagebackref,breaklinks,colorlinks,allcolors=cvprblue]{hyperref}
\newcommand{\modelname}[1]{MobileVLA-R1}
\newcommand{\datasetname}[1]{MobileVLA-CoT}
\usepackage{pifont}
\usepackage{multicol,multirow}
\usepackage{caption}
\newcommand{\cmark}{\ding{51}} %
\newcommand{\xmark}{\ding{55}} %
\def\paperID{10938} %
\def\confName{CVPR}
\def\confYear{2026}

\title{\raisebox{-0.5em}{\includegraphics[height=1.5em]{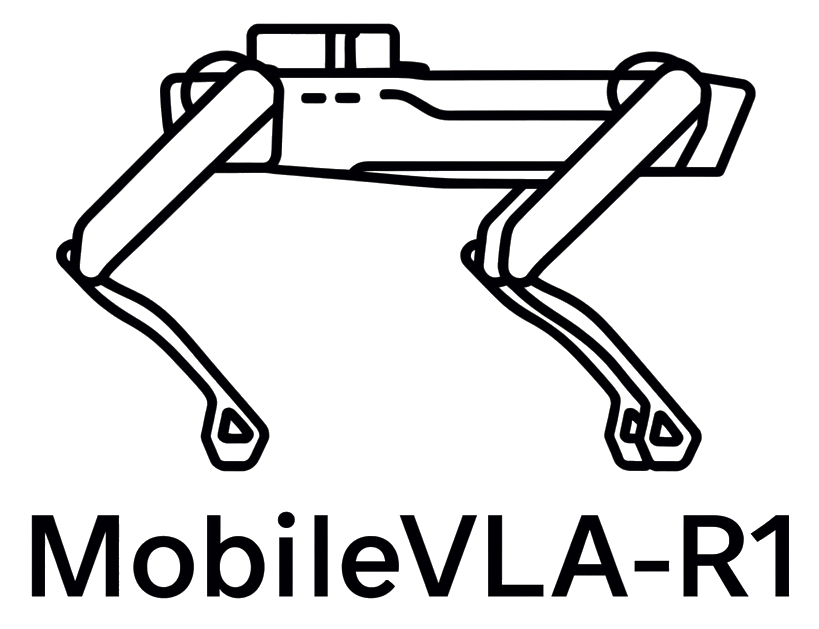}}~\modelname{}: Reinforcing Vision-Language-Action for Mobile Robots}

\author{
    \textbf{Ting Huang}$^{*}$ \quad
    \textbf{Dongjian Li}$^{*}$ \quad
    \textbf{Rui Yang}$^{*}$ \quad
    \textbf{Zeyu Zhang}$^{*\dag}$ \quad
    \textbf{Zida Yang} \quad
    \textbf{Hao Tang}$^{\ddag}$ \vspace{0.1cm}\\
    Peking University\\
    \small $^*$Equal contribution. $^\dag$Project lead.
    $^\ddag$Corresponding author: bjdxtanghao@gmail.com.
}

\begin{document}

\makeatletter
\let\@oldmaketitle\@maketitle%
\renewcommand{\@maketitle}{\@oldmaketitle%
\vspace{0.2cm}

{\centering
\vspace{-0.35cm}
\includegraphics[width=\linewidth]{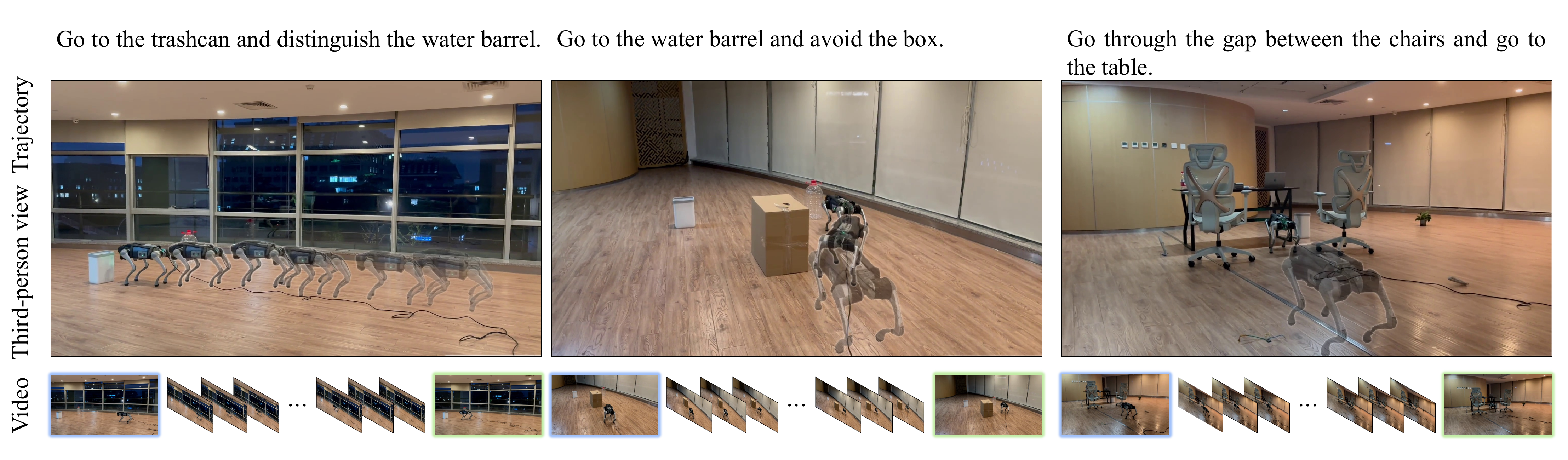}\par
\captionof{figure}{\textbf{Real-world demonstration of \modelname{}.} Upon receiving natural-language instructions, \modelname{} processes RGB video streams through a vision–language model to perform spatial reasoning and generate continuous locomotion commands, enabling the quadruped robot to accomplish complex tasks in real-world environments.}
\label{fig:main}}
\bigskip}%
\makeatother

\maketitle

\begin{abstract}
Grounding natural-language instructions into continuous control for quadruped robots remains a fundamental challenge in vision language action.
Existing methods struggle to bridge high-level semantic reasoning and low-level actuation, leading to unstable grounding and weak generalization in the real-world.
To address these issues, we present \modelname{}, a unified vision–language–action framework that enables explicit reasoning and continuous control for quadruped robots.
We construct \datasetname{}, a large-scale dataset of multi-granularity CoT for embodied trajectories, providing structured reasoning supervision for alignment.
Built upon this foundation, we introduce a two-stage training paradigm that combines supervised CoT alignment with GRPO reinforcement learning to enhance reasoning consistency, control stability, and long-horizon execution.
Extensive evaluations on VLN and VLA tasks demonstrate superior performance over strong baselines, with approximately a 5\% improvement.
Real-world deployment on a quadruped robot validates robust performance in complex environments.
Code: \url{https://github.com/AIGeeksGroup/MobileVLA-R1}.
Website: \url{https://aigeeksgroup.github.io/MobileVLA-R1}.
\end{abstract}

\section{Introduction}
\label{sec:intro}

Vision-language navigation (VLN) and embodied manipulation are fundamental capabilities for intelligent robotic systems, enabling agents to ground natural-language instructions into perception, reasoning, and continuous control in real-world environments~\cite{8578485,Wang2018ReinforcedCM,Chaplot2020ObjectGN,Driess2023PaLMEAE,rt2,saycan2022arxiv,Huang2022LanguageMA}.
Such competencies are essential for service robotics, assistive systems, and general-purpose autonomous agents operating in human-centric spaces, where natural-language interaction and reliable execution are required.

Recent advances in multimodal foundation models have significantly improved perception and language understanding, motivating research on extending vision-language models (VLMs) to embodied tasks.
Despite this progress, two major gaps still limit current approaches.
The first lies in bridging semantic reasoning and motor control: direct language-to-action mapping often leads to weak interpretability and unstable grounding.
The second involves a lack of transparent reasoning structures: methods relying on latent intermediate embeddings achieve stability but obscure semantic logic, limiting compositional reasoning and error traceability.

Our motivation is to build an embodied foundation model that can bridge high-level semantic reasoning and low-level motor execution, enabling interpretable planning and robust control across diverse environments.

To this end, we introduce \modelname{}, a hierarchical vision language action (VLA) framework designed for quadruped robots.
Rather than predicting motor commands directly from language, \modelname{} generates structured Chain-of-Thought (CoT) action plans conditioned on multimodal observations and then translates them into continuous control commands through an action decoder.
This reasoning-then-execution design enhances interpretability, improves grounding stability, and enables flexible adaptation across diverse environments.
The training process follows a two-stage paradigm that first aligns reasoning ability through supervised learning on CoT-annotated embodied data and then refines action grounding and execution fidelity via Group Relative Policy Optimization (GRPO) reinforcement learning.

We evaluate \modelname{} across two complementary embodied AI benchmarks: R2R-CE and RxR-CE for language-guided navigation, and the QUARD dataset for quadruped control and manipulation.
On VLN-CE, \modelname{} achieves state-of-the-art performance with an average improvement of 5\% in success rate over strong baselines, while on QUARD it consistently outperforms recent vision–language–action models across all six control tasks, demonstrating enhanced reasoning-aligned control.
Real-world deployment on the Unitree Go2 further confirms robust performance under cluttered and partially observable conditions.

In this work, we make three key contributions:
\begin{itemize}
    \item We propose \modelname{}, a hierarchical vision language action framework that explicitly connects semantic reasoning and motor control through Chain-of-Thought generation and continuous quadruped execution, addressing the core semantic–control gap.
    \item We design a two-stage training framework that integrates supervised CoT alignment with GRPO reinforcement learning, improving reasoning consistency, control robustness, and long-horizon stability.
    \item We construct \datasetname{}, a multi-granularity CoT dataset for embodied trajectories, and demonstrate measurable gains on embodied AI benchmarks, achieving approximately 5\% performance improvement and reliable deployment on the Unitree Go2 platform.
\end{itemize}

\begin{figure*}[t]
    \centering
    \includegraphics[width=0.9\linewidth]{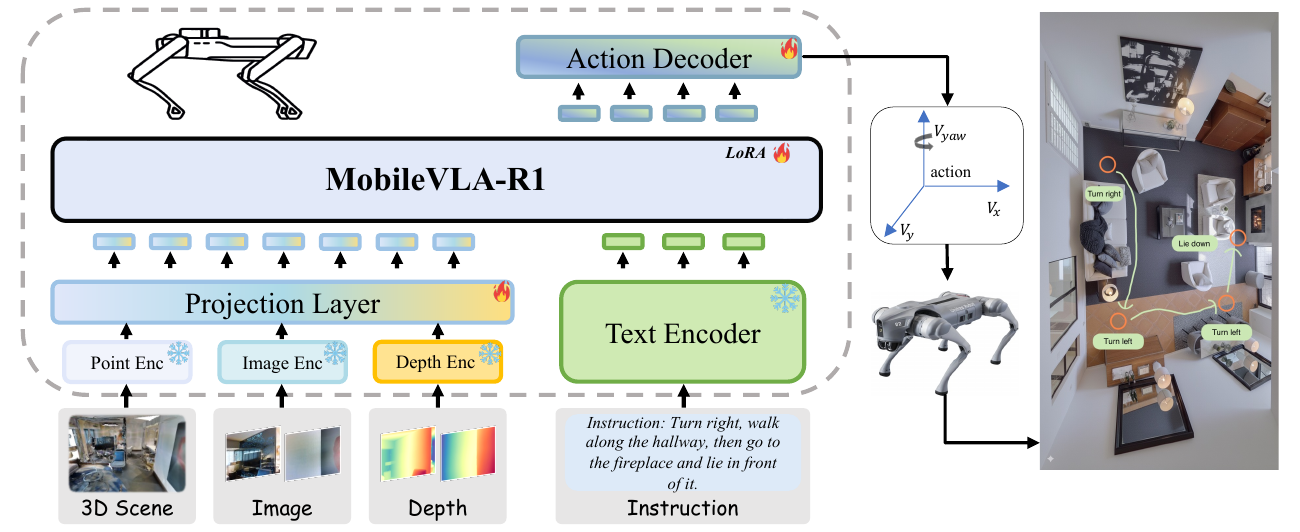}\par
    \caption{
    \textbf{Architecture of \modelname{}.} \modelname{} is an end-to-end framework that integrates natural-language instructions with multimodal perception. It processes RGB, depth, and point cloud observations together with textual commands to generate continuous locomotion actions, enabling mobile robots to follow complex instructions and adapt to diverse environments in real time.
    }
    \label{fig:model}
    \vspace{-0.4cm}
\end{figure*}

\section{Related Work}
\noindent\textbf{Mobile robot navigation and manipulation.}
Vision language navigation (VLN) studies how agents follow natural-language instructions in 3D environments.
R2R~\cite{8578485} established the benchmark with human-annotated indoor trajectories, and RxR~\cite{ku2020room} extended it with multilingual, longer, and semantically richer descriptions.
Early approaches relied on sequence-to-sequence models~\cite{fried2018speaker,ma2019selfmonitoring}, while later works introduced attention and transformer-based architectures~\cite{021681dfbef84fc796af7876b1110a76,zhu2020vision,chen2021history} to improve cross-modal grounding and long-horizon reasoning.
With large-scale pretrained VLMs~\cite{huang20253d,huang20253dcoca,huang2025dc}, recent VLN systems~\cite{Hong_2021_CVPR,hao2020prevalent,Qi2021TheRT,10.1609/icaps.v34i1.31506,Yu2023L3MVNLL,vlnr1,zhang2024uninavid,zhu2025mtu,liu2025nav} further enhance instruction following and generalization across diverse scenes.

In parallel, language-conditioned robot manipulation has shown that natural language can directly guide physical actions~\cite{song2025hazards,song2025maniplvm,ye2025vla,liu2025evovla}.
Systems such as SayTap~\cite{Tang2023SayTapLT}, RT-2~\cite{rt2}, OpenVLA~\cite{kim24openvla}, and generalist embodied frameworks like Octo~\cite{octonav2025} demonstrate robust instruction-conditioned locomotion and manipulation in real-world settings.
These developments collectively point toward unified embodied agents capable of grounding language in both navigation and manipulation behaviors.

\noindent\textbf{Reinforcement learning on LLMs.}
Transformer-based language models~\cite{Vaswani2017AttentionIA,radford2019language,devlin2019bert} exhibit strong generalization through large-scale pretraining, while RLHF~\cite{Ouyang2022TrainingLM} further improves instruction following.
Chain-of-thought (CoT) prompting~\cite{10.5555/3600270.3602070} enhances multi-step reasoning, and recent models such as DeepSeek-R1 combine CoT with GRPO~\cite{Shao2024DeepSeekMathPT} to improve long-horizon reasoning. CoT has also been extended to spatial and embodied settings~\cite{Chen2025IntegratingCF}, suggesting its potential for grounded decision making.

In embodied AI, reinforcement learning is increasingly used to align multimodal perception, reasoning, and control. RL-based frameworks like RLVR-World~\cite{rlvrworld2025} improve semantic consistency and robustness in real-world tasks, while multimodal reasoning systems such as MM-HELIX~\cite{mmhelix2025}, Octo~\cite{octonav2025}, and OpenVLA~\cite{kim24openvla} demonstrate the promise of integrating high-level reasoning with continuous control.
Overall, these directions reveal a growing convergence between CoT reasoning, reinforcement learning, and embodied policy optimization, motivating unified frameworks that connect abstract reasoning with low-level robotic execution.

\section{Datasets}
\label{sec:datasets}

\subsection{Public Datasets}
R2R~\cite{8578485} provides instruction–trajectory pairs in realistic Matterport3D~\cite{matterport3D2017} indoor scenes and serves as a canonical benchmark for instruction-following navigation.
RxR~\cite{rxr2020} extends this setting with multilingual supervision and dense temporal grounding, offering longer and more instruction-dense descriptions that enrich spatial semantics.
To supply continuous control and manipulation signals on legged platforms, we additionally use QUARD~\cite{ding2025quar}, a quadruped dataset encompassing both locomotion and manipulation episodes in diverse environments.
Together, these sources provide complementary supervision for language grounding, spatial reasoning, and low-level action execution.

\begin{table}[t]
\caption{
\textbf{Statistics of public VLN and embodied-control datasets used for synthesizing \datasetname{}.}
``Nav.'' indicates navigation capability, ``Emb-Ctl.'' indicates embodied control capability, and ``CoT Anno.'' indicates whether the dataset provides chain-of-thought annotations.}
\label{tab:dataset_comparison}
\centering
\resizebox{\linewidth}{!}{
    \begin{tabular}{lcccr}
        \toprule
        Dataset                              & Nav. & Emb-Ctl. & CoT Anno. & Scale \\
        \midrule
        R2R~\cite{8578485}                   & \cmark & \xmark & \xmark & 50K \\
        RxR~\cite{rxr2020}                   & \cmark & \xmark & \xmark & 58 \\
        QUARD~\cite{ding2025quar}            & \cmark & \cmark & \xmark & 262K \\
        \midrule
        \textbf{MobileVLA-CoT-Episode}       & \xmark & \cmark & \cmark & 18K  \\
        \textbf{MobileVLA-CoT-Step}          & \xmark & \cmark & \cmark & 78K \\
        \textbf{MobileVLA-CoT-Nav}           & \cmark & \xmark & \cmark & 38K \\
        \bottomrule
    \end{tabular}
}
\vspace{-0.4cm}
\end{table}

\subsection{Synthetic Dataset}
Building on R2R, RxR, and QUARD, we synthesize \textbf{\datasetname{}}, a large-scale chain-of-thought corpus aligned with multimodal observations and control targets.
Unlike datasets that only specify final goals, \datasetname{} explicitly records the reasoning process that connects instructions to actions, enabling interpretable and step-by-step decision traces.
It contains episode-level CoT, which summarizes task outcomes and high-level execution plans; step-level CoT, which specifies the next action and its rationale; and navigation-level CoT, which connects global instructions with multi-step reasoning across the trajectory.
In total, the corpus includes 18K episode-level samples, 78K step-level samples, and a navigation-focused subset, MobileVLA-CoT-Nav, with 38K CoT annotations, as summarized in \tableautorefname~\ref{tab:dataset_comparison}.

\subsection{CoT Data Engine}
\begin{figure*}[t]
    \centering
    \includegraphics[width=\linewidth]{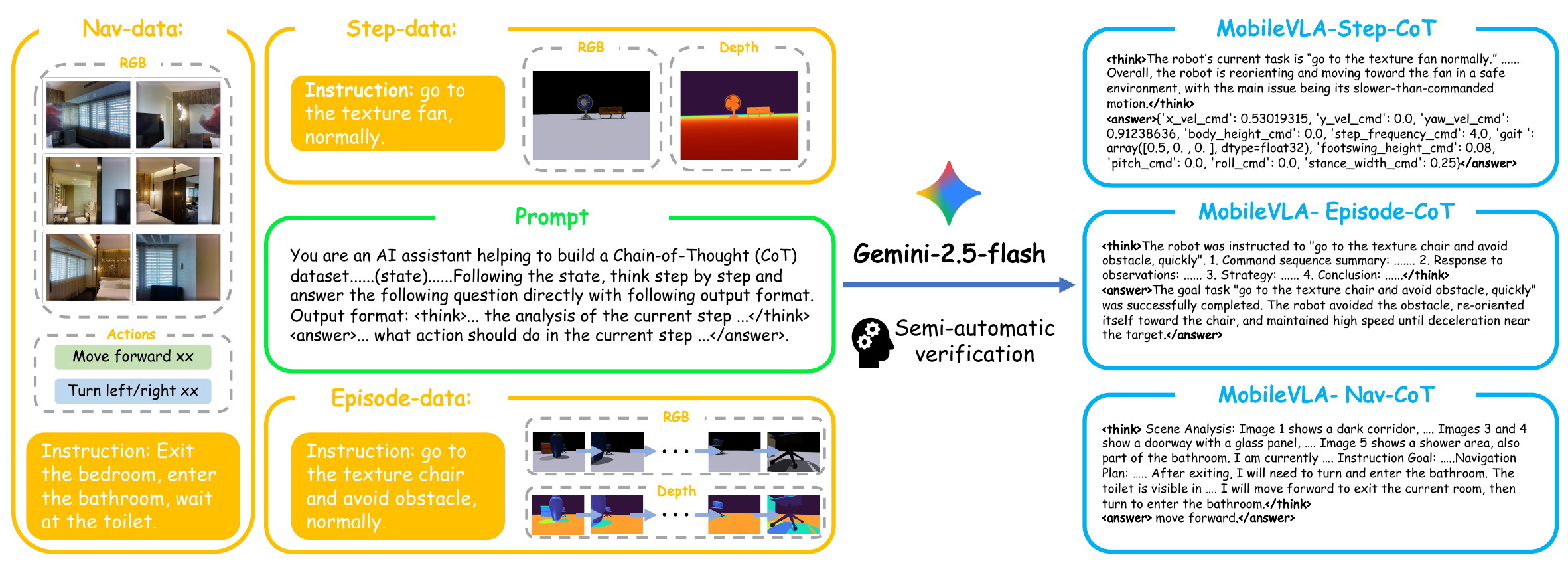}
    \caption{
    \textbf{CoT Data Engine.} We construct the \datasetname{} by defining navigation and step-level instructions, integrating RGB–Depth visual inputs, and specifying structured reasoning prompts. These inputs are fed into Gemini-2.5-Flash, which generates multi-granularity Chain-of-Thought (CoT) annotations with corresponding action outputs.}
    \label{fig:cot}
    \vspace{-0.4cm}
\end{figure*}

As illustrated in \figureautorefname~\ref{fig:cot}, we design a unified CoT data engine to synthesize multi-granularity reasoning traces for navigation and embodied control tasks.
The engine takes paired RGB–Depth observations, natural-language instructions, and corresponding state–action histories as inputs and leverages the reasoning capability of a large multimodal model, Gemini-2.5-Flash~\cite{gemini2025}, guided by structured prompt templates.
Each prompt describes the current environment, specifies the task format, and instructs the model to reason before answering.
The model first generates the reasoning process enclosed in \texttt{<think>...</think>} tags, followed by the corresponding control command or action in \texttt{<answer>...</answer>} format.

To cover different reasoning granularities, we construct three complementary datasets: MobileVLA-CoT-Nav provides long-horizon navigation-level reasoning across full trajectories, MobileVLA-CoT-Step captures local control reasoning and next-action prediction, and MobileVLA-CoT-Episode summarizes trajectory-level outcomes and high-level strategies.
A semi-automatic verification pipeline, combining rule-based filtering and manual inspection, is applied to remove malformed or unsafe annotations before training; additional details on data generation and quality control are provided in the \textit{Supp. Mat.} \sectionautorefname~\ref{sec:supp_data_quality}.
Through this process, the engine generates high-quality, reasoning-aligned embodied trajectories that serve as the foundation for supervised CoT alignment in \modelname{}, enabling structured reasoning and reliable control grounding before reinforcement learning.

\section{The Proposed Method}

\subsection{Overview}
\begin{figure*}[t]
    \centering
    \includegraphics[width=0.9\linewidth]{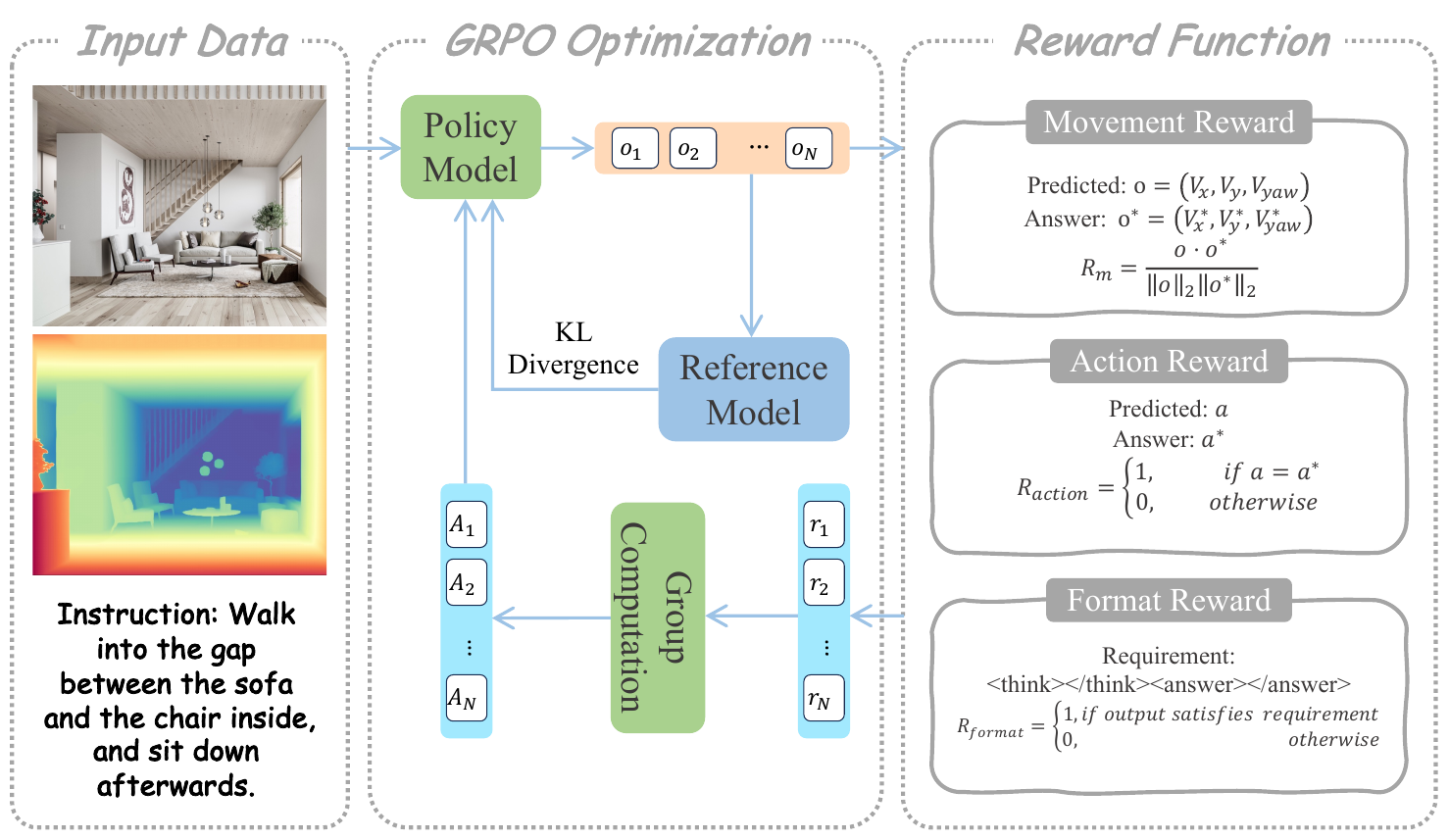}
    \caption{
    \textbf{The pipeline of RL policy.} The model generates $N$ responses from a given input, rewards are then computed for each response. After normalizing and clipping, these rewards are conflated with a KL-divergence term, which prevents the model from over-updating, to update the policy.
    }
    \label{fig:rl}
    \vspace{-0.4cm}
\end{figure*}

\modelname{} follows a hierarchical reasoning–execution paradigm composed of two sequential training stages.
Built upon the \datasetname{} dataset (Section~\ref{sec:datasets}), the framework first performs supervised fine-tuning (SFT) on CoT-annotated embodied data to establish structured reasoning and semantic alignment between visual observations and textual commands.
The second stage further refines action grounding and continuous control through Group Relative Policy Optimization (GRPO)~\cite{Shao2024DeepSeekMathPT}, as depicted in \figureautorefname~\ref{fig:rl}.

For architecture, as illustrated in \figureautorefname~\ref{fig:model}, \modelname{} follows the LLaVA~\cite{Liu2023VisualIT} design, adopting a unified multimodal perception front-end that integrates RGB, depth, and point cloud inputs.
We initialize the model from NaVILA~\cite{cheng2025navila} and jointly train it with SFT and GRPO, achieving a tight coupling between explicit reasoning and continuous control within a unified vision language action framework.

\noindent\textbf{Problem setups.}
The objective of \modelname{} is to generate a continuous sequence of actions for closed-loop quadruped control, conditioned on multimodal perception and natural-language instructions.
At each timestep $t$, the agent receives an observation $s_t = \{x_t^{\text{rgb}}, x_t^{\text{depth}}, x_t^{\text{point}}\}$ together with an instruction $i \in I$, and predicts an action $a_t \in A$ according to a policy $\pi_\theta(a_t \mid s_t, i)$.
Formally, the mapping can be expressed as:
\begin{equation}
\resizebox{0.9\hsize}{!}{$
    f_\theta: (X^{\text{rgb}}, X^{\text{depth}}, X^{\text{point}}) \times I \rightarrow A,
    \quad a_t = [V_x, V_y, \omega_{yaw}, \alpha],
$}
\end{equation}
where $V_x$ and $V_y$ denote the translational velocities along the $x$- and $y$-axes, $\omega_{yaw}$ represents the angular velocity around the yaw axis, and $\alpha$ is a discrete high-level action sampled from a predefined action set.
This formulation unifies continuous locomotion control and discrete embodied behaviors under a reasoning-conditioned policy, enabling interpretable and robust quadruped operation in real-world environments.

\subsection{Cold Start Stage}
Recent works such as DeepSeek-R1~\cite{DeepSeekAI2025DeepSeekR1IR} highlight the potential of reinforcement learning for reasoning; yet direct end-to-end optimization often causes instability in multimodal settings.
To mitigate this, we introduce a cold-start stage based on supervised fine-tuning (SFT) to align the output format and initialize reasoning before reinforcement learning.

This stage involves two steps.
First, the model is fine-tuned on the MobileVLA-CoT-Episode and MobileVLA-CoT-Nav datasets to learn structured reasoning in the format \texttt{<think>...</think><answer>...</answer>}, thereby improving coherence in long-horizon tasks.
Then, it is further trained on a 10K-sample subset of MobileVLA-CoT-Step, where answers contain executable velocity and action commands.
These are deterministically parsed into control signals, enabling a smooth transition from language reasoning to physical actuation.

\subsection{Reinforcement Learning}

To further enhance reasoning-driven action generation, we employ the GRPO framework~\cite{Shao2024DeepSeekMathPT}, which has recently shown remarkable success in DeepSeek-R1~\cite{DeepSeekAI2025DeepSeekR1IR}.
Unlike traditional gradient-based reinforcement learning algorithms such as Proximal Policy Optimization (PPO)~\cite{Schulman2017ProximalPO}, GRPO focuses on reward alignment within groups of sampled responses, allowing for more stable and efficient learning from comparative feedback rather than absolute reward magnitudes.
Its objective is to iteratively refine the model’s policy through group-wise reward optimization, thereby improving both reasoning coherence and action precision.

\noindent\textbf{Policy sample.} Given an input pair $(x, i)$, where $x$ encodes a multi-modal scene representation composed of RGB images, depth maps, and point cloud features, and $i$ represents a natural language instruction, the policy $\pi_\theta$ generates $N$ responses $\{o_1, o_2, \cdots, o_N\}$ for each sample.
Each response is evaluated using a set of reward functions that encourage accurate, executable, and well-formatted outputs.

\noindent\textbf{Movement reward.} This reward evaluates the consistency between the predicted and ground-truth motion directions, encouraging smooth and stable trajectories.
The policy outputs a control vector $\mathbf{o} = (V_x, V_y, V_{yaw}, a)$, where the first three components represent continuous velocities.
Given the ground truth $\mathbf{o^*} = (V_x^*, V_y^*, V_{yaw}^*, a^*)$, the movement reward is computed as the cosine similarity between the two velocity vectors:
\begin{equation}
R_m =
\frac{\mathbf{o'}^\top \mathbf{o'^*}}
{\lVert \mathbf{o'} \rVert_2 \, \lVert \mathbf{o'^*} \rVert_2},
\end{equation}
where $\mathbf{o'} = (V_x, V_y, V_{yaw})$.
This formulation promotes directionally aligned and dynamically consistent control, while discrete actions are supervised separately by the action reward.

\noindent\textbf{Action reward.} This reward supervises discrete control behaviors by checking whether the predicted action $a$ matches the ground-truth label $a^*$.
It provides a simple yet effective signal for aligning high-level decisions with target actions:
\begin{equation}
R_{action} =
\begin{cases}
1, & \text{if } a = a^*\\
0, & \text{otherwise}
\end{cases}.
\end{equation}
Such a binary formulation offers clear feedback on action correctness and complements the continuous guidance from the movement reward.

\noindent\textbf{Format reward.} To ensure structural consistency, the format reward $R_{format}$ checks whether the model output strictly follows the reasoning–execution template \texttt{<think>...</think><answer>...</answer>}.
Formally,
\begin{equation}
    R_{format} =
    \begin{cases}
        1, & \text{if output satisfies format}\\
        0, & \text{otherwise}
    \end{cases}.
\end{equation}
This constraint guarantees machine-parseable outputs and enforces consistent mapping from reasoning traces to executable control commands.

\noindent\textbf{Policy update.} Given $N$ responses sampled from the current policy $\pi_\theta$, we compute individual rewards $\mathbf{r} = (r_1, r_2, \dots, r_N)$ and advantages
\begin{equation}
A_i = r_i - \frac{1}{N} \sum_{j=1}^{N} r_j,
\end{equation}
which are normalized as $\hat{A}_i = \frac{A_i}{\sigma_A + \epsilon}$ for stability.
The policy is then optimized using the clipped GRPO objective:
\begin{equation}
\resizebox{0.9\hsize}{!}{$
\begin{split}
& J_{\mathrm{GRPO}}(\theta) = 
\mathbb{E}_{j=1}^G \Biggl[
    \min \Biggl(
        \frac{\pi_\theta(o_j \mid I)}{\pi_{\theta_{\mathrm{old}}}(o_j \mid I)} \hat{A}_j, \\
        & \operatorname{clip}\!\left(
            \frac{\pi_\theta(o_j \mid I)}{\pi_{\theta_{\mathrm{old}}}(o_j \mid I)},
            1 - \epsilon,\; 1 + \epsilon
        \right) \hat{A}_j
    \Biggr) 
    - \beta \cdot D_{\mathrm{KL}}\!\left(\pi_\theta \,\|\, \pi_{\text{ref}}\right)
\Biggr]
\end{split}
$}
\end{equation}
where $I=(x,i)$ is the multimodal observation–instruction pair, $o_j$ is the $j$-th generated output, and $\pi_{\text{ref}}$ denotes the frozen reference policy.

\begin{table*}[t]
\centering
\caption{\textbf{Comparison with SOTA methods on the Val-Unseen split of VLN-CE~\cite{krantz_vlnce_2020,ku2020room}.} ``Pano'' stands for panorama and ``Odo'' stands for odometry. $^*$indicates methods using the waypoint predictor~\cite{Hong2022BridgingTG}. \modelname{} outperforms all methods that do not rely on simulator pre-trained waypoint predictors, even when those methods leverage additional inputs such as depth, panoramic views, and odometry.}
\label{tab:vlnce}
\resizebox{0.88\textwidth}{!}{
\begin{tabular}{lcccclcccclcccc}
    \toprule
    \multirow{2}[2]{*}{Method} & \multicolumn{4}{c}{Observation} & & \multicolumn{4}{c}{R2R-CE Val-Unseen} & & \multicolumn{4}{c}{RxR-CE Val-Unseen} \\
    \cmidrule(lr){2-5} \cmidrule(lr){7-10} \cmidrule(lr){12-15}
    & S.RGB & Pano. & Depth & Odom. & & NE$\downarrow$ & OS$\uparrow$ & SR$\uparrow$ & SPL$\uparrow$ & & NE$\downarrow$ & SR$\uparrow$ & SPL$\uparrow$ & nDTW$\uparrow$ \\
    \midrule
    CMA$^{*}$~\cite{Hong2022BridgingTG}         & \xmark & \cmark & \cmark & \cmark & & 6.20 & 52.0 & 41.0 & 36.0 & & 8.76 & 26.5 & 22.1 & 47.0 \\
    Sim2Sim$^{*}$~\cite{krantz2022sim}          & \xmark & \cmark & \cmark & \cmark & & 6.07 & 52.0 & 43.0 & 36.0 & & - & - & - & - \\
    GridMM$^{*}$~\cite{wang2023gridmm}          & \xmark & \cmark & \cmark & \cmark & & 5.11 & 61.0 & 49.0 & 41.0 & & - & - & - & - \\
    Ego$^{2}$-Map$^{*}$~\cite{hong2023learning} & \xmark & \cmark & \cmark & \cmark & & 5.54 & 56.0 & 47.0 & 41.0 & & - & - & - & - \\
    DreamWalker$^{*}$~\cite{wang2023dreamwalker}& \xmark & \cmark & \cmark & \cmark & & 5.53 & 59.0 & 49.0 & 44.0 & & - & - & - & - \\
    Reborn$^{*}$~\cite{an20221st}               & \xmark & \cmark & \cmark & \cmark & & 5.40 & 57.0 & 50.0 & 46.0 & & 5.98 & 48.6 & 42.0 & 63.3 \\
    ETPNav$^{*}$~\cite{an2024etpnav}            & \xmark & \cmark & \cmark & \cmark & & 4.71 & 65.0 & 57.0 & 49.0 & & 5.64 & 54.7 & 44.8 & 61.9 \\
    HNR$^{*}$~\cite{wang2024lookahead}          & \xmark & \cmark & \cmark & \cmark & & 4.42 & 67.0 & 61.0 & 51.0 & & 5.50 & 56.3 & 46.7 & 63.5 \\
    \midrule
    AG-CMTP~\cite{chen2021topological}          & \xmark & \cmark & \cmark & \cmark & & 7.90 & 39.0 & 23.0 & 19.0 & & - & - & - & - \\
    R2R-CMTP~\cite{chen2021topological}         & \xmark & \cmark & \cmark & \cmark & & 7.90 & 38.0 & 26.0 & 22.0 & & - & - & - & - \\
    InstructNav~\cite{long2024instructnavzeroshotgenericinstruction} & \xmark & \cmark & \cmark & \cmark & & 6.89 & - & 31.0 & 24.0 & & - & - & - & - \\
    LAW~\cite{raychaudhuri2021language}         & \cmark & \xmark & \cmark & \cmark & & 6.83 & 44.0 & 35.0 & 31.0 & & 10.90 & 8.0 & 8.0 & 38.0 \\
    CM2~\cite{georgakis2022cross}               & \cmark & \xmark & \cmark & \cmark & & 7.02 & 41.0 & 34.0 & 27.0 & & - & - & - & - \\
    WS-MGMap~\cite{chen2022weakly}              & \cmark & \xmark & \cmark & \cmark & & 6.28 & 47.0 & 38.0 & 34.0 & & - & - & - & - \\
    AO-Planner~\cite{chen2024affordances}       & \xmark & \cmark & \cmark & \xmark & & 5.55 & 59.0 & 47.0 & 33.0 & & 7.06 & 43.3 & 30.5 & 50.1 \\
    Seq2Seq~\cite{krantz2020beyond}             & \cmark & \xmark & \cmark & \xmark & & 7.77 & 37.0 & 25.0 & 22.0 & & 12.10 & 13.9 & 11.9 & 30.8 \\
    CMA~\cite{krantz2020beyond}                 & \cmark & \xmark & \cmark & \xmark & & 7.37 & 40.0 & 32.0 & 30.0 & & - & - & - & - \\
    NaVid~\cite{zhang2024navid}                 & \cmark & \xmark & \xmark & \xmark & & 5.47 & 49.0 & 37.0 & 35.0 & & - & - & - & - \\
    Uni-NaVid~\cite{zhang2024uninavid}          & \cmark & \xmark & \xmark & \xmark & & 5.58 & 53.5 & 47.0 & 42.7 & & 6.24 & 48.7 & 40.9 & - \\
    NaVILA~\cite{cheng2025navila}               & \cmark & \xmark & \xmark & \xmark & & 5.22 & 62.5 & 54.0 & 49.0 & & 6.77 & 49.3 & 44.0 & 58.8 \\
    VLN-R1~\cite{vlnr1}                         & \cmark & \xmark & \xmark & \xmark & & 7.00 & 41.2 & 30.2 & 21.8 & & 9.10 & 22.7 & 17.6 & -    \\
    OctoNav~\cite{octonav2025}                  & \cmark & \xmark & \xmark & \xmark & & -    & 42.9 & 37.1 & 33.6 & & -    & -    & -    & -    \\
    StreamVLN~\cite{streamvln}                  & \cmark & \xmark & \xmark & \xmark & & 4.98 & 64.2 & 56.9 & 51.9 & & 6.22 & 52.9 & 46.0 & 61.9 \\
    CorrectNav~\cite{correctnav2025}            & \cmark & \xmark & \xmark & \xmark & & 4.24 & 67.5 & 65.1 & 62.3 & & 4.09 & 69.3 & 63.3 & 75.2 \\
    \midrule
    \textbf{\modelname{} (Ours)}                & \cmark & \xmark & \cmark & \xmark & & \textbf{4.05} & \textbf{69.7} & \textbf{68.3} & \textbf{65.2} & & \textbf{3.92} & \textbf{71.5} & \textbf{66.8} & \textbf{76.1} \\
    \bottomrule
\end{tabular}
}
\vspace{-0.4cm}
\end{table*}

\section{Experiments}

\noindent\textbf{Benchmarks.}
To comprehensively evaluate both the navigation and embodied control capabilities of \modelname{}, we adopt two complementary settings.
For high-level navigation reasoning, we evaluate \modelname{} on the VLN-CE benchmarks~\cite{krantz_vlnce_2020, ku2020room}, which provide continuous 3D environments reconstructed from Matterport3D~\cite{matterport3D2017}.
Compared with R2R~\cite{8578485} and RxR~\cite{rxr2020}, VLN-CE supports continuous interaction with the environment, posing greater challenges for long-horizon spatial reasoning and low-latency control.
We follow standard protocols and report results on the \textit{val-unseen} splits of VLN-CE-R2R and VLN-CE-RxR to measure navigation accuracy, success rate, and trajectory efficiency.

To further assess the low-level actuation and reasoning-aligned control ability, we conduct action-only evaluations on the QUARD dataset~\cite{ding2025quar}, which contains diverse quadruped locomotion and manipulation tasks.
While VLN-CE focuses on high-level navigation under linguistic instructions, QUARD emphasizes the precise execution of continuous control commands and discrete behaviors.

\noindent\textbf{Evaluation metrics.}
For the VLN-CE benchmarks, we adopt standard navigation metrics, including navigation error (NE), oracle success rate (OS), success rate (SR), success-weighted path length (SPL), and normalized dynamic time warping (nDTW).
These metrics collectively evaluate the agent’s spatial reasoning accuracy, trajectory efficiency, and overall instruction-following success.

For embodied control evaluation on QUARD~\cite{ding2025quar}, we follow the official protocol and report the average success rate across six quadruped control tasks.
Each value represents the success ratio over 25 evaluation episodes, and tasks are grouped into three difficulty levels (``Easy'', ``Medium'', ``Hard'') based on motion complexity and interaction difficulty.

\noindent\textbf{Network architecture.} We initialize \modelname{} from the pretrained NaVILA~\cite{cheng2025navila} to leverage its embodied vision–language alignment.
Since NaVILA was originally designed for RGB-only perception, we extend its architecture by introducing additional encoders for depth~\cite{wu2025stereoadapter} and point cloud modalities.
Specifically, we incorporate DepthAnything V2~\cite{Yang2024DepthAV} as the depth encoder and a Point Transformer v3~\cite{wu2024ptv3} for 3D geometric representation.
The multi-modal features from RGB, depth, and point inputs are fused through a lightweight projection module before being aligned with the LLaMA3-8B language backbone.

\noindent\textbf{Parameter efficient tuning.} To enable scalable fine-tuning under limited computation, we adopt LoRA~\cite{Hu2021LoRALA} as the primary parameter-efficient tuning strategy.
LoRA modules are inserted into the projection and attention layers of the LLaMA3-8B backbone, while the vision encoders remain frozen.
This design preserves pretrained multimodal representations from NaVILA while allowing efficient adaptation to embodied reasoning and control tasks.

We first perform supervised alignment on approximately 20K steps using the full MobileVLA-CoT-Episode and MobileVLA-CoT-Nav datasets, along with a small subset of the MobileVLA-CoT-Step dataset.
The goal is to establish structured reasoning and decision-making abilities.
We adopt the LoRA configuration with $r=16$, $\alpha=32$, and a cosine learning-rate scheduler.
Training is conducted on 4$\times$H20 (96 GB) GPUs with a learning rate of $2\times10^{-4}$, using the AdamW optimizer.
Following SFT, we refine the model via Group Relative Policy Optimization (GRPO)~\cite{Shao2024DeepSeekMathPT} to improve reasoning-to-action consistency and execution stability.
In each update, the model generates 8 responses per sample, and 5 samples are used for each optimization step.
Training runs for 1K steps on a single H20 (96 GB) GPU with $\beta=0.04$, $clip\_param=0.2$, and a learning rate of $1\times10^{-6}$.
The policy is optimized with AdamW, using KL regularization to stabilize updates and maintain alignment with the SFT reference model.

\begin{table}[t!]
\caption{\textbf{Overall performance on the QUARD benchmark.} We evaluate \modelname{} and baselines on six quadruped control tasks from QUARD~\cite{ding2025quar}, grouped by difficulty into ``Easy'', ``Medium'', and ``Hard''.
Each value denotes the average success rate over 25 evaluation episodes.
Our model achieves consistently higher performance across both locomotion and manipulation tasks, demonstrating robust reasoning-aligned control in challenging embodied settings.}
\label{tab:exp_baseline}
\centering
\resizebox{\linewidth}{!}{
\begin{tabular}{l|c|c|cccc|c}
\toprule
\multirow{2}{*}{Method} & Easy & Medium & \multicolumn{4}{c|}{Hard} & \multirow{2}{*}{Average} \\
\cmidrule{2-2} \cmidrule{3-3} \cmidrule{4-7}
 & Distinguish & Go to & Go avoid & Go through & Crawl & Unload &  \\
\midrule
CLIP~\cite{radford2021learning}     & 0.44 & 0.43 & 0.45 & 0.19 & 0.00 & 0.00 & 0.25 \\
VC-1~\cite{vc2023}                  & 0.46 & 0.43 & 0.45 & 0.31 & 0.00 & 0.00 & 0.28 \\
QUART~\cite{ding2024quarvla}        & 0.66 & 0.60 & 0.53 & 0.41 & 0.32 & 0.12 & 0.44 \\
MoRE~\cite{zhao2025more}            & 0.82 & 0.80 & 0.59 & 0.57 & 0.49 & 0.33 & 0.60 \\
\midrule
\textbf{\modelname{}} & \textbf{0.92} & \textbf{0.89} & \textbf{0.71} & \textbf{0.65} & \textbf{0.58} & \textbf{0.44} & \textbf{0.73} \\
\bottomrule
\end{tabular}
}
\vspace{-0.4cm}
\end{table}

\subsection{Main Results}
\noindent\textbf{Vision language navigation.} As shown in Table~\ref{tab:vlnce}, \modelname{} consistently outperforms prior methods on both R2R-CE~\cite{krantz_vlnce_2020} and RxR-CE~\cite{ku2020room} benchmarks.
It achieves higher success rates and SPL scores with lower navigation errors, demonstrating superior reasoning alignment and generalization across unseen environments.

\noindent\textbf{Quadruped control and manipulation.} As shown in Table~\ref{tab:exp_baseline}, \modelname{} achieves the best overall performance on the QUARD benchmark.
It consistently surpasses strong baselines such as QUART~\cite{ding2024quarvla} and MoRE~\cite{zhao2025more} across both locomotion and manipulation tasks, demonstrating stable reasoning-to-control grounding and robust execution under complex conditions.

\begin{figure*}[t]
    \centering
    \includegraphics[width=\linewidth]{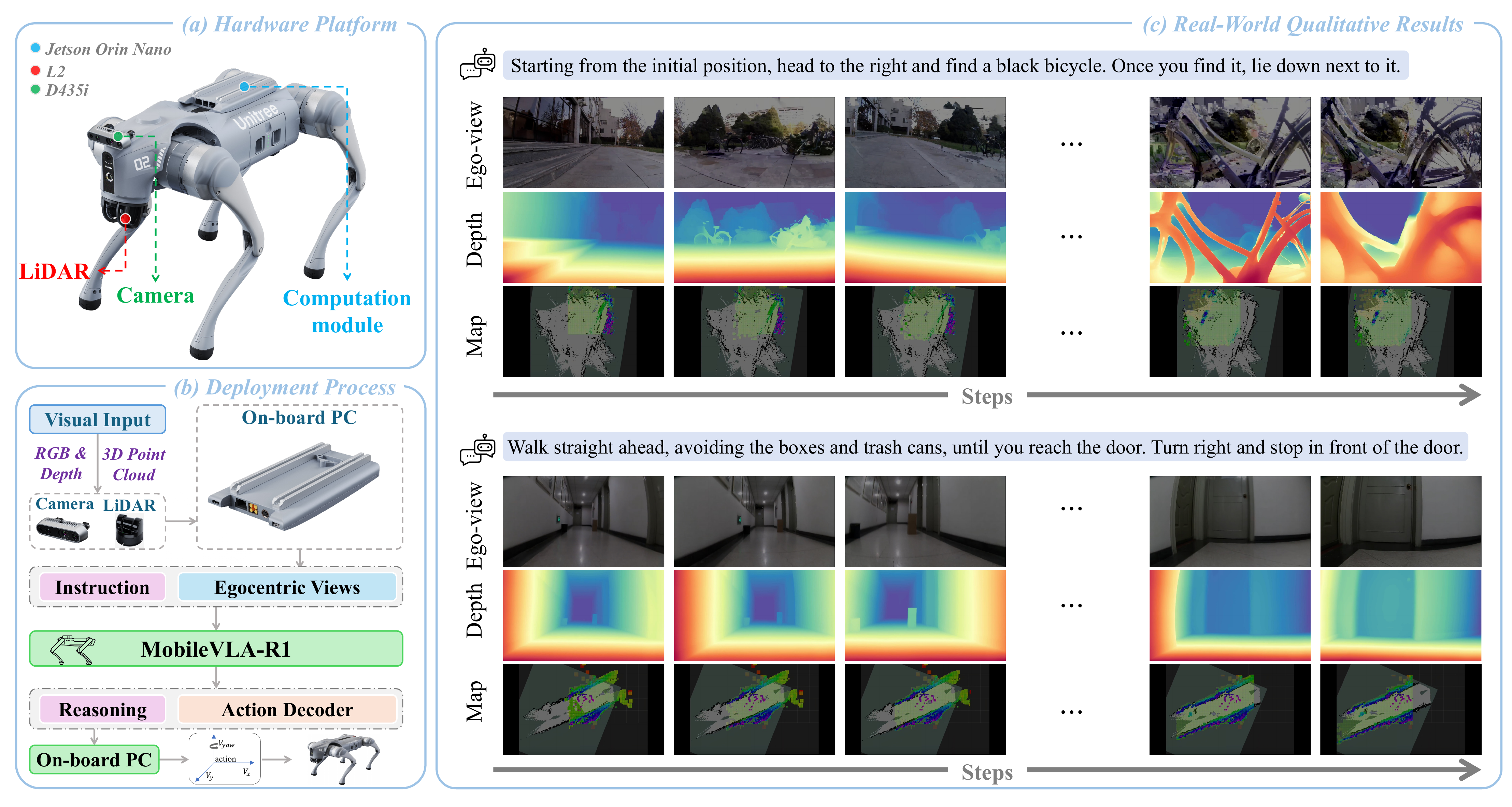}
    \caption{
    \textbf{(a) Hardware platform:} the Unitree Go2 quadruped robot is equipped with a Jetson Orin Nano (on-board PC) as the computation module, an L2 LiDAR for 3D environment perception, and an Intel RealSense D435i RGB-D camera for visual sensing.
    \textbf{(b) Deployment process:} RGB–Depth and 3D point cloud data are transmitted to \modelname{}, which performs multimodal reasoning and action generation. The resulting velocity and motion commands are sent back to the on-board PC for real-time execution on the robot.
    \textbf{(c) Real-World qualitative results:} \modelname{} effectively integrates RGB, depth, and map observations to follow long-horizon language instructions with coherent spatial reasoning.
    }
    \label{fig:real_world}
    \vspace{-0.4cm}
\end{figure*}

\subsection{Real World Evaluation}

\noindent\textbf{Robot type settings.}
As shown in Fig.~\ref{fig:real_world} (a), we use the Unitree Go2 quadruped robot as the physical platform for real-world evaluation.
The robot is equipped with an L2 LiDAR for 3D environmental perception, an RGB-D camera for panoramic visual sensing, and an NVIDIA Jetson Orin computation module for onboard inference.
As illustrated in Fig.~\ref{fig:real_world} (b), the sensory inputs are synchronized and fused into a unified point–depth representation, which is processed by \modelname{} in real time.
When onboard resources are limited, the inference pipeline operates in a hybrid mode: the Go2 transmits sensory data to a remote H20 server, where \modelname{} performs reasoning and sends continuous velocity commands back for execution.
This closed-loop design ensures low-latency control while maintaining consistent coordination between perception, reasoning, and actuation.

\noindent\textbf{Scene setup and task types.}
To evaluate generalization under diverse real-world conditions, we conduct experiments in three representative environments: \texttt{Workspace}, \texttt{Corridor}, and \texttt{Outdoor}.
For each, we define two levels of instruction complexity: \textit{Simple} tasks consist of one or two short navigation or manipulation commands, whereas \textit{Complex} tasks require multi-step reasoning and long-horizon spatial planning.
This setup enables a comprehensive assessment of robustness, reasoning-to-action grounding, and adaptability to environmental diversity.

\noindent\textbf{Quantitative real-world evaluation.}
\modelname{} consistently outperforms prior models, including NaVILA~\cite{cheng2025navila} and GPT-4o~\cite{openai2024gpt4o}, across Workspace, Corridor, and Outdoor environments (Table~\ref{tab:real_eval}).
It achieves higher success rates and lower navigation errors under both simple and complex instructions, demonstrating robust reasoning-to-control alignment and stable execution in real-world scenarios.

\begin{table}[!t]
\centering
\caption{\textbf{Real-world experiments on quadruped (Unitree Go2)} conducted in different environments (\texttt{Workspace}, \texttt{Corridor}, and \texttt{Outdoor}). Simple and Complex refer to simple and complex instruction-following tasks, respectively.}
\label{tab:real_eval}
\resizebox{\linewidth}{!}{
\begin{tabular}{l cccc cccc cccc}
\toprule
\multirow{3}{*}{Method} & \multicolumn{4}{c}{\texttt{Workspace}} & \multicolumn{4}{c}{\texttt{Corridor}} & \multicolumn{4}{c}{\texttt{Outdoor}} \\
\cmidrule(lr){2-5} \cmidrule(lr){6-9} \cmidrule(lr){10-13}
& \multicolumn{2}{c}{Simple} & \multicolumn{2}{c}{Complex} & \multicolumn{2}{c}{Simple} & \multicolumn{2}{c}{Complex} & \multicolumn{2}{c}{Simple} & \multicolumn{2}{c}{Complex} \\
\cmidrule(lr){2-3} \cmidrule(lr){4-5} \cmidrule(lr){6-7} \cmidrule(lr){8-9} \cmidrule(lr){10-11} \cmidrule(lr){12-13}
& NE$\downarrow$ & SR$\uparrow$ & NE$\downarrow$ & SR$\uparrow$ & NE$\downarrow$ & SR$\uparrow$ & NE$\downarrow$ & SR$\uparrow$ & NE$\downarrow$ & SR$\uparrow$ & NE$\downarrow$ & SR$\uparrow$ \\ 
\midrule
GPT-4o~\cite{openai2024gpt4o}    & 2.01 & 0.67 & 2.38 & 0.33 & 1.49 & 0.53 & 3.00 & 0.00 & - & 0.67 & - & 0.50 \\
NaVILA~\cite{cheng2025navila}    & 1.29 & 0.83 & 1.76 & 0.80 & 1.15 & 0.89 & 1.76 & 0.67 & - & 0.91 & - & 0.83 \\ \midrule
\textbf{\modelname{}} & \textbf{1.03} & \textbf{0.93} & \textbf{1.23} & \textbf{0.91} & \textbf{0.96} & \textbf{1.00} & \textbf{1.23} & \textbf{0.86} & - & \textbf{1.00} & - & \textbf{0.96} \\
\bottomrule
\end{tabular}
}
\vspace{-0.4cm}
\end{table}

\noindent\textbf{Qualitative real-world evaluation.}
In Fig.~\ref{fig:main} and Fig.~\ref{fig:real_world} (c), \modelname{} demonstrates smooth and coherent locomotion across various real-world scenes.
It accurately follows complex natural-language commands, maintaining stable motion and spatial consistency under dynamic and cluttered conditions.
These results highlight the effectiveness of our reasoning-aligned framework in bridging high-level intent and low-level control for robust quadruped execution.
More qualitative visualizations will be included in the \textit{Supp. Mat.} \sectionautorefname~\ref{sec:visualization}.

\subsection{Ablation Study}
\begin{table}[t]
\centering
\caption{\textbf{Reward decomposition on R2R-CE Val-Unseen.} \cmark denotes inclusion of the reward.
The first row is the \textbf{SFT-only} baseline, while subsequent rows apply GRPO with different subsets of movement ($R_m$), action ($R_{action}$), and format ($R_{format}$) rewards.
Removing any term degrades SR and SPL, and using all three yields the best navigation success and path efficiency.}
\resizebox{0.6\linewidth}{!}{
    \begin{tabular}{ccccc}
    \toprule
    $R_{m}$ & $R_{action}$ & $R_{format}$ & SR$\uparrow$ & SPL$\uparrow$ \\
    \midrule
    \xmark & \xmark & \xmark & 58.0 & 53.2 \\ \midrule
    \cmark & \xmark & \xmark & 60.7 & 55.5 \\
    \xmark & \cmark & \xmark & 61.9 & 56.8 \\
    \xmark & \xmark & \cmark & 59.6 & 54.7 \\
    \midrule
    \cmark & \cmark & \xmark & 64.5 & 60.2 \\
    \cmark & \xmark & \cmark & 63.4 & 59.1 \\
    \xmark & \cmark & \cmark & 65.2 & 61.0 \\
    \midrule
    \cmark & \cmark & \cmark & \textbf{68.3} & \textbf{65.2} \\
    \bottomrule
    \end{tabular}}
\label{tab:ablation_rewards}
\vspace{-0.4cm}
\end{table}

\noindent\textbf{Effect of reward design.} To verify the effectiveness of our reward formulation in GRPO training, we ablate the three reward components: the movement reward $R_{m}$, the action reward $R_{action}$, and the format reward $R_{format}$.
Table~\ref{tab:ablation_rewards} reports the performance on the R2R-CE Val-Unseen split.
Removing any rewards causes a noticeable degradation in navigation success and trajectory efficiency.
Without $R_{m}$, the model exhibits unstable locomotion and oscillatory motion.
Excluding $R_{action}$ weakens high-level decision consistency, while dropping $R_{format}$ leads to malformed or unstructured outputs that cannot be reliably parsed into executable commands.
Combining all three rewards yields the highest success rate and SPL, validating that they provide complementary optimization signals for reasoning-aligned and control-stable policy learning.

\noindent\textbf{Additional ablation.} We further conduct incremental modality encoder ablations to analyze the contribution of each perception stream in the \textit{Supp. Mat.} \sectionautorefname~\ref{sec:ablastudy}.

\section{Conclusion}

In this work, we introduce \modelname{}, a unified vision–language–action framework that bridges high-level reasoning and low-level control for quadruped robots.
By decoupling structured chain-of-thought reasoning from continuous motor execution, our model achieves interpretable decision-making and robust control across diverse environments.
The two-stage training paradigm, which combines supervised CoT alignment with GRPO-based reinforcement learning, effectively enhances reasoning consistency, control stability, and long-horizon execution.
Extensive experiments on the VLN-CE and QUARD benchmarks, as well as real-world deployments on the Unitree Go2 robot, demonstrate the superior performance and adaptability of \modelname{} compared to existing methods.
These results highlight the effectiveness of integrating structured reasoning with continuous control, advancing the development of generalizable embodied agents.
\clearpage
{
    \small

\input{main.bbl}
}

\input{X_suppl}

\end{document}

%% file: X_suppl.tex
\clearpage
\setcounter{page}{1}
\maketitlesupplementary

\section{More Ablation Study}
\label{sec:ablastudy}

\noindent\textbf{Effect of multimodal perception.}
We perform an incremental modality encoder ablation to examine the contributions of each sensory modality.
Starting from the text and image encoder, we progressively add depth and point cloud encoders to the model.
As shown in Table~\ref{tab:modal_increment}, each additional modality leads to consistent improvements in navigation success (SR) and trajectory efficiency (SPL) on both R2R-CE and RxR-CE.
Depth cues enhance local geometric understanding and obstacle awareness, while point cloud features further strengthen 3D spatial reasoning and path planning.
The full multimodal configuration achieves the best overall performance, demonstrating that rich 3D perception substantially benefits reasoning-grounded embodied navigation.

\begin{table}[ht]
\centering
\caption{\textbf{Incremental modality encoder ablation on VLN-CE benchmarks.}
Starting from the text and image encoder, we incrementally add depth and point cloud encoders.
Each additional modality consistently improves navigation success and trajectory efficiency on both R2R-CE and RxR-CE.
The full configuration \modelname{} achieves the best performance, demonstrating the benefit of rich 3D perception for reasoning-grounded navigation.}
\resizebox{0.9\linewidth}{!}{
\begin{tabular}{lcccc}
\toprule
\multirow{2}{*}{Setting} & \multicolumn{2}{c}{R2R-CE} & \multicolumn{2}{c}{RxR-CE} \\
\cmidrule(lr){2-3}\cmidrule(lr){4-5}
 & SR$\uparrow$ & SPL$\uparrow$ & SR$\uparrow$ & SPL$\uparrow$ \\
\midrule
Text \& Image Encoder        & 62.5 & 59.0 & 66.0 & 61.5 \\
\ + Depth Encoder            & 66.0 & 62.0 & 69.0 & 64.0 \\
\ + Point Encoder            & 67.2 & 63.5 & 70.2 & 65.2 \\ \midrule
\textbf{\modelname{} (Ours)} & \textbf{68.3} & \textbf{65.2} & \textbf{71.5} & \textbf{66.8} \\
\bottomrule
\end{tabular}}
\label{tab:modal_increment}
\end{table}

\section{CoT Data Generation and Quality Control}
\label{sec:supp_data_quality}

\noindent\textbf{Avoiding data leakage.}
We generate CoT annotations only for the official training splits of R2R, RxR, and QUARD, and do not synthesize data on any validation or test episodes.
All quantitative results are obtained on the original benchmarks using their human-provided instructions and evaluation protocols, and the teacher model (Gemini-2.5-Flash) is never queried with ground-truth answers for the evaluated trajectories.

\noindent\textbf{Semi-automatic verification pipeline.}
To control annotation quality, we adopt a semi-automatic verification pipeline.
We first apply rule-based parsing and regular-expression checks to filter out malformed outputs that violate the \texttt{<think>...</think>} and \texttt{<answer>...</answer>} formats, contain unparsable control commands, or propose actions outside the QUARD action space.
The remaining samples are then manually reviewed on a randomly sampled subset from each of the three CoT subsets.
During this pass, two authors independently check whether (i) the reasoning is logically consistent with the visual observations and instructions, (ii) the proposed action is feasible given the state-action history, and (iii) there are no obvious safety issues, such as collisions.
Samples that fail any of these criteria are either removed or corrected, and we additionally spot-check long-horizon and cluttered episodes to identify systematic failure modes.
In practice, the most common errors are overly generic reasoning or slightly mis-scaled velocity commands, which our filtering removes before training.

\section{Test-Time Efficiency and Practical Considerations}
\label{sec:efficiency}

Test-time efficiency is critical for deploying embodied models on real quadruped platforms.
In our current implementation, \modelname{} runs on a remote H20 GPU server and communicates with the robot via a wired connection through a local computer.
The vision-language backbone has 8B parameters, which allows for strong reasoning capabilities but also introduces non-trivial latency.
On this setup, one forward pass of \modelname{} takes about $10$ seconds per decision step, and the end-to-end closed loop (including perception I/O and network transmission) requires roughly $15$ seconds per control step.

Despite this latency, the current system is adequate for our experimental settings, which focus on indoor navigation and short-range manipulation in quasi-static environments, where the robot moves cautiously and the scene changes slowly.
However, we acknowledge that such latency is too high for highly dynamic scenarios, fast locomotion, or fine-grained real-time interaction, and we view this as a key limitation of the present prototype.

There are several practical trade-offs and mitigation strategies.
First, the action decoder in \modelname{} already converts CoT outputs into continuous commands over a short horizon, allowing the robot to execute multiple low-level steps per high-level reasoning cycle.
This suggests a regime where CoT-based planning runs at a lower frequency (e.g., when the instruction or scene changes), while lightweight on-board controllers track the issued velocity commands at a higher frequency.
Second, the visual encoders and language backbone can, in principle, be replaced by smaller architectures or compressed via distillation and quantization to trade a modest amount of accuracy for substantial reductions in latency and memory footprint.
Third, caching perception features and sharing CoT prefixes across consecutive steps can amortize the cost of long-horizon reasoning.

A systematic exploration of these design choices is beyond the scope of this work, but our results highlight that achieving strong CoT-based reasoning on quadruped robots is currently possible, even under relatively high latency, and point toward model compression and hierarchical control as promising directions for practical deployment.

\section{Visualization}
\label{sec:visualization}

We provide qualitative visualizations in both simulator and real-world settings to illustrate the perception–reasoning–action process of \modelname{} across diverse environments.

\begin{figure*}[ht]
    \centering
    \includegraphics[width=\linewidth]{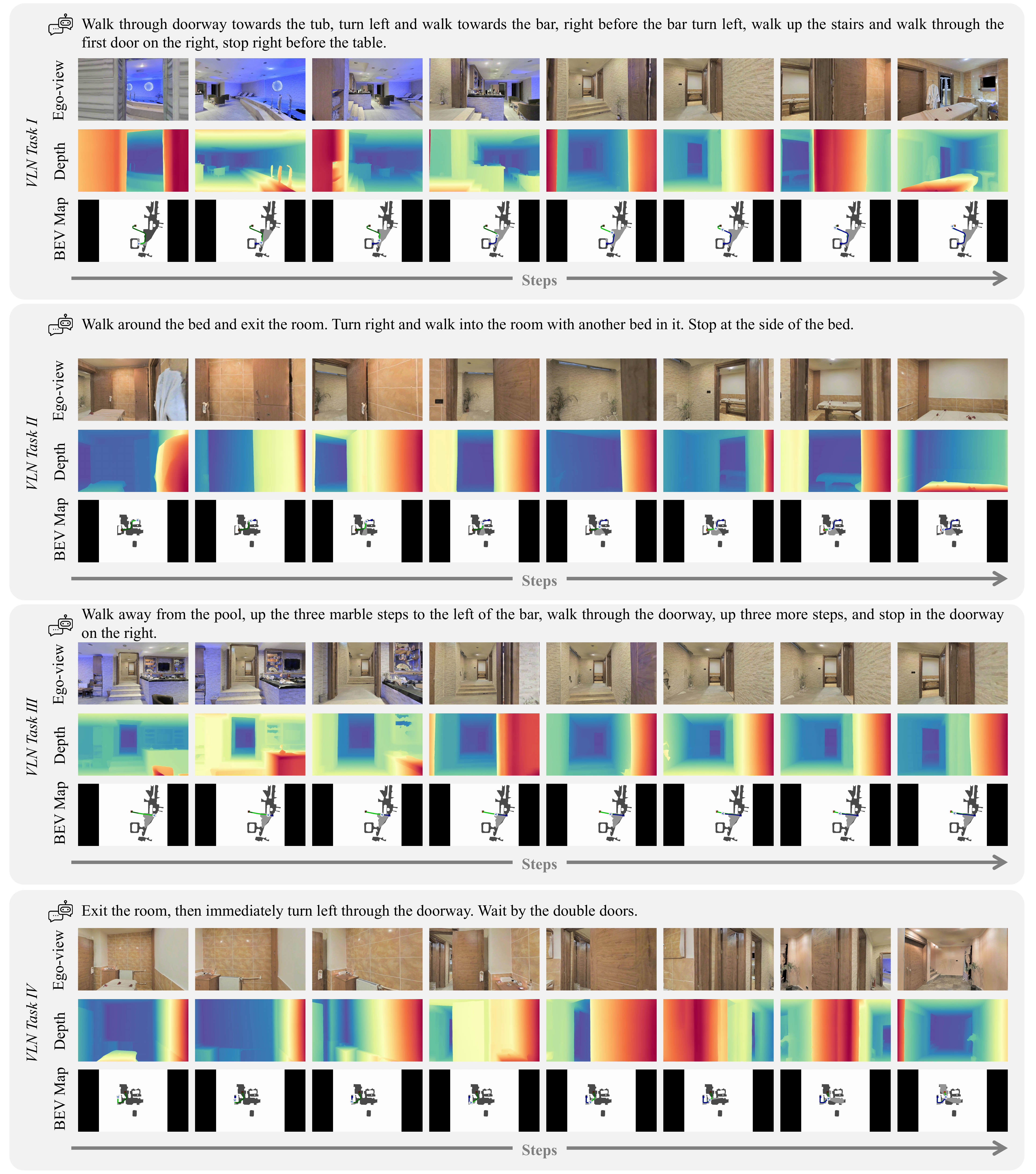}
    \caption{\textbf{Qualitative visualization on R2R-CE.} \modelname{} executes English navigation instructions in the simulator.
    }
    \label{fig:simulator_vln_1}
\end{figure*}

\begin{figure*}[ht]
    \centering
    \includegraphics[width=\linewidth]{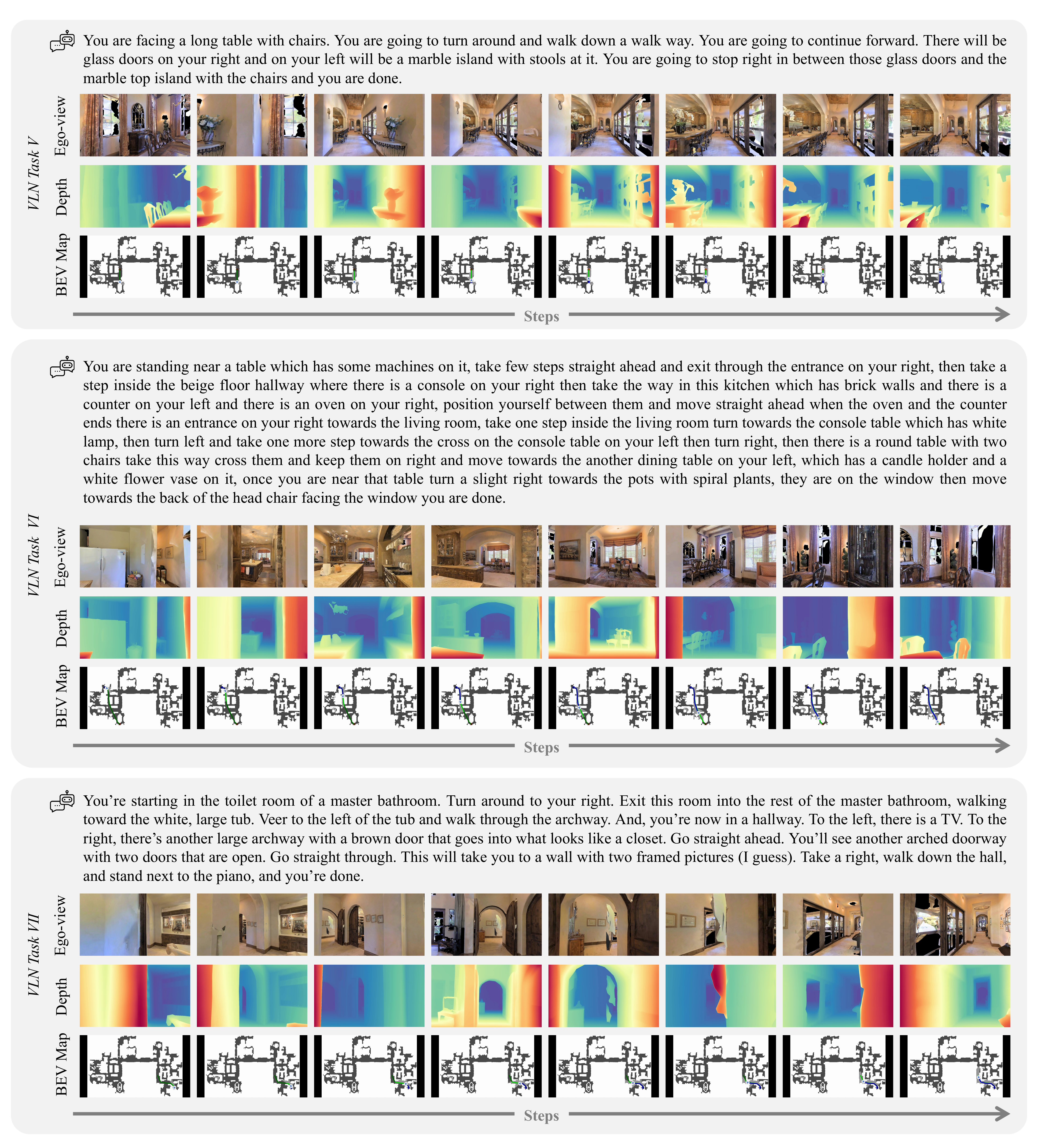}
    \caption{\textbf{Qualitative visualization on RxR-CE.} \modelname{} follows multilingual instructions with complex spatial semantics, maintaining coherent reasoning and stable motion across unseen environments.
    }
    \label{fig:simulator_vln_2}
\end{figure*}

As shown in \figureautorefname~\ref{fig:simulator_vln_1} and \figureautorefname~\ref{fig:simulator_vln_2}, the robot receives natural-language instructions and performs step-by-step reasoning before executing continuous actions in the R2R-CE and RxR-CE validation splits.
Each trajectory demonstrates coherent visual perception, spatial reasoning, and action grounding under unseen scenes and linguistic variations, highlighting the model’s robust cross-lingual understanding and spatial consistency.

\begin{figure*}[ht]
    \centering
    \includegraphics[width=\linewidth]{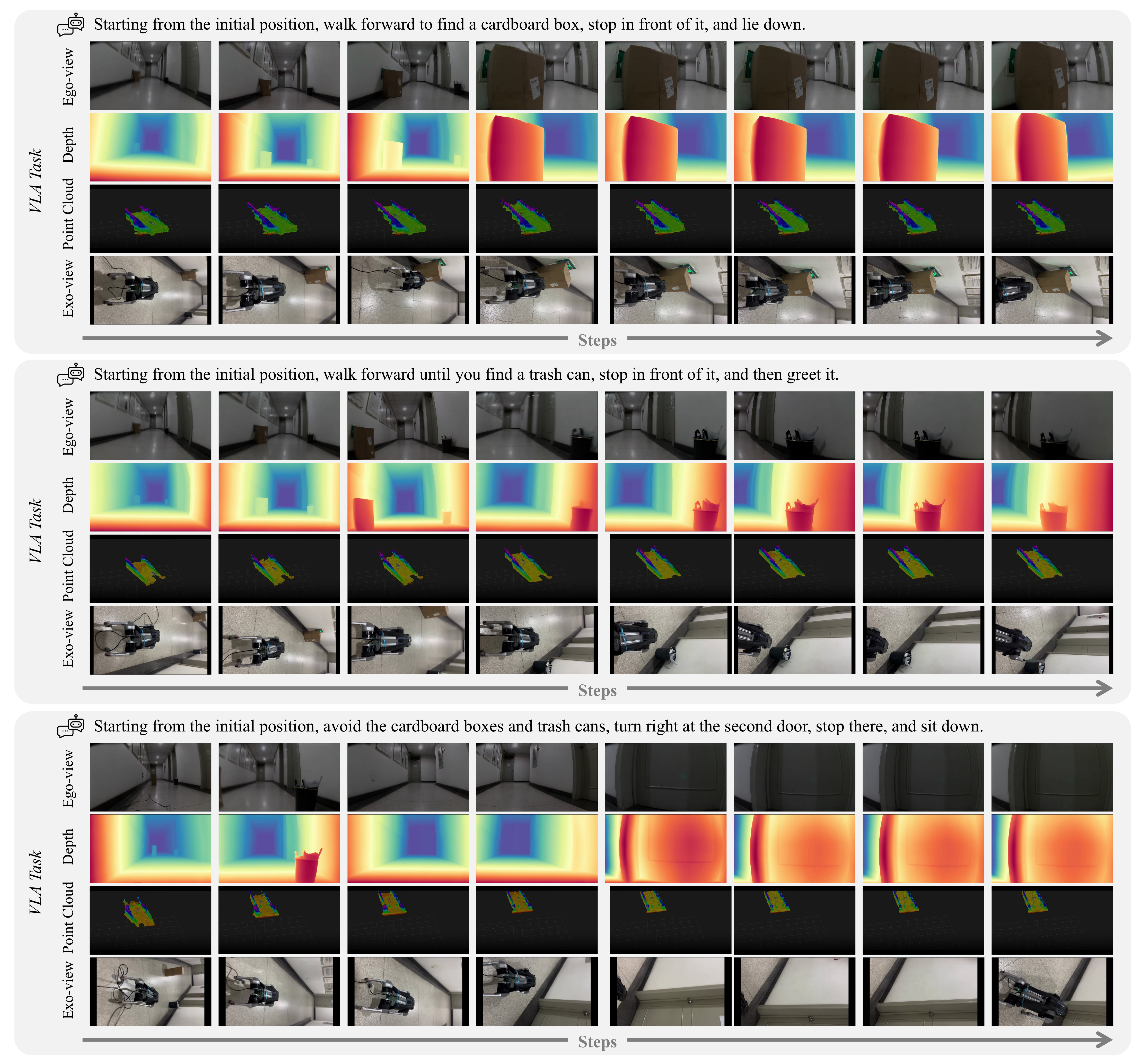}
    \caption{\textbf{Real-world qualitative results.} The robot demonstrates spatially grounded perception–reasoning–action behaviors, such as approaching targets, interacting with objects, and avoiding obstacles in realistic indoor environments.
    }
    \label{fig:realworld_vis1}
\end{figure*}

\begin{figure*}[ht]
    \centering
    \includegraphics[width=\linewidth]{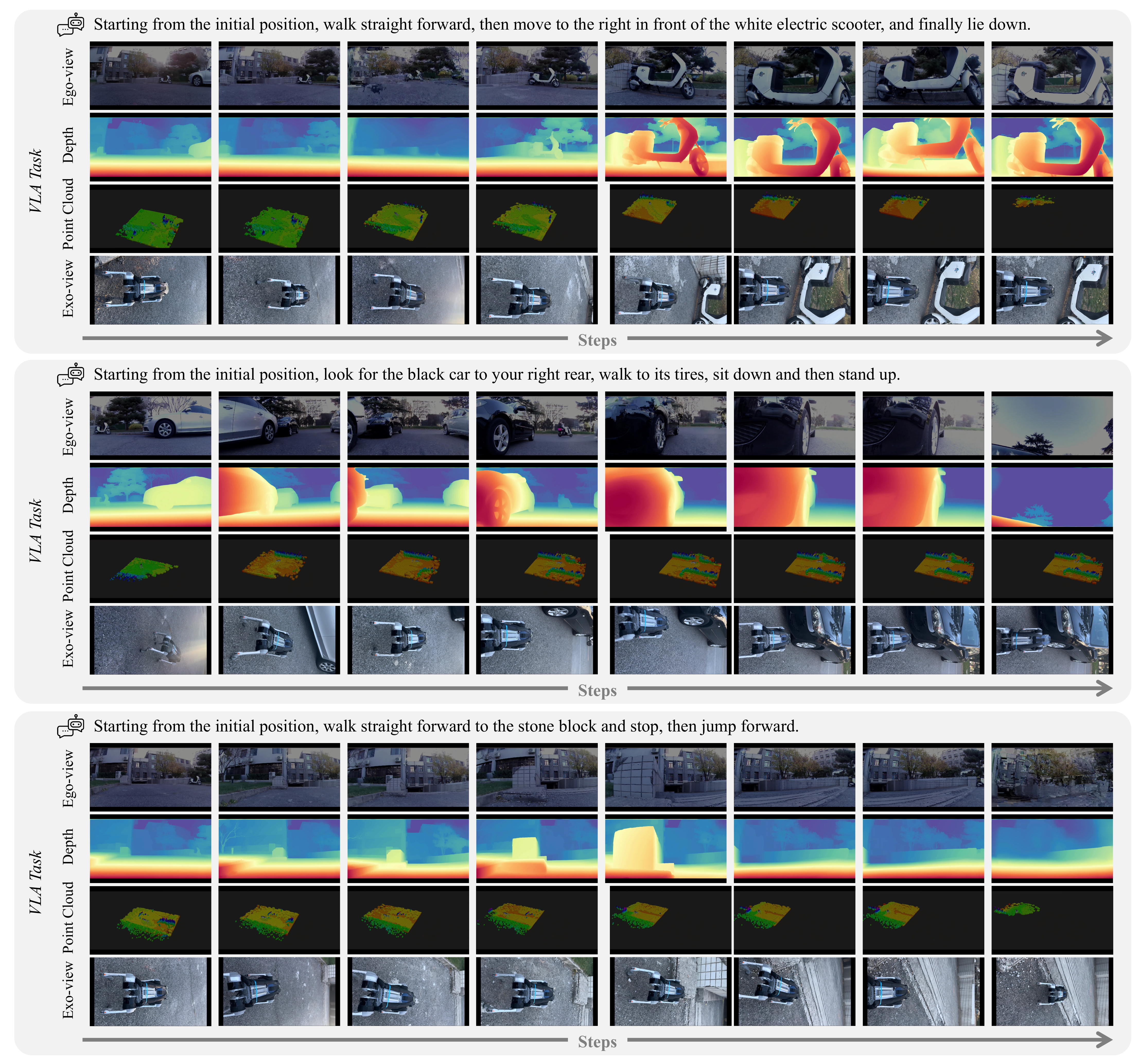}
    \caption{\textbf{Additional outdoor real-world qualitative results.} \modelname{} exhibits fine-grained obstacle avoidance, target-aware spatial reasoning, and high-precision locomotion control in diverse real-world outdoor scenes.
    }
    \label{fig:realworld_vis3}
\end{figure*}

Finally, \figureautorefname~\ref{fig:realworld_vis1} - \figureautorefname~\ref{fig:realworld_vis3} present real-world deployments on the Unitree Go2 robot.
Under complex indoor and outdoor environments, \modelname{} accurately follows long-horizon commands, maintaining stable motion and continuous spatial grounding.
These results validate that \modelname{} generalizes from simulated training to physical environments with robust reasoning-aligned control.

\begin{figure*}[ht]
    \centering
    \includegraphics[width=\linewidth]{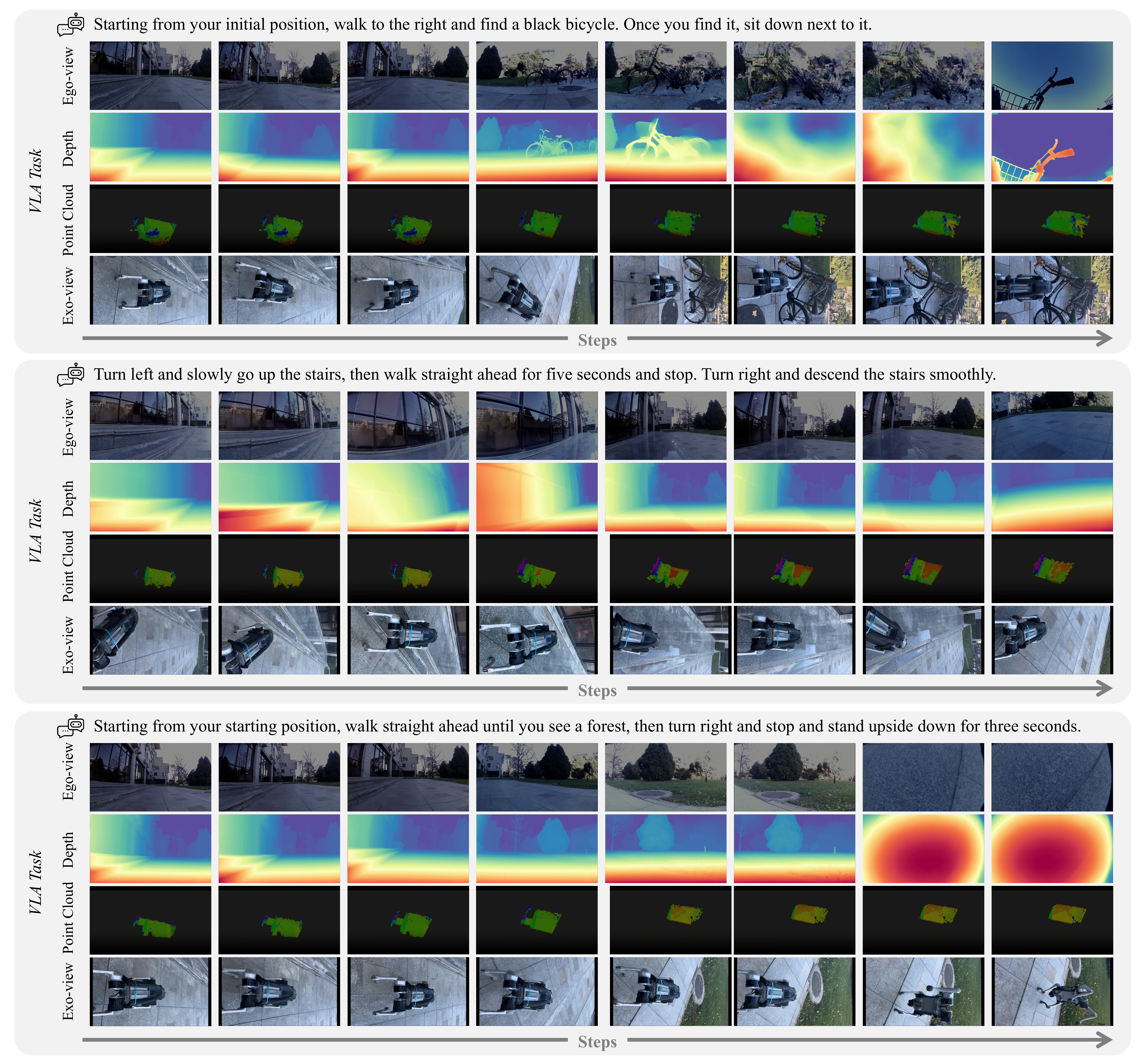}
    \caption{\textbf{Outdoor real-world qualitative results.} \modelname{} delivers highly reliable long-horizon reasoning and control in diverse outdoor environments, showcasing its strong real-world generalization.
    }
    \label{fig:realworld_vis2}
\end{figure*}

\section{Limitation and Future Work}

\noindent\textbf{Limitations.}
While \modelname{} achieves strong performance in both simulation and real-world environments, several limitations remain that highlight directions for future research.
First, the current framework mainly focuses on navigation and basic manipulation, where the action decoder maps high-level language instructions to a fixed set of predefined behaviors.
This design limits the system’s ability to cope with highly dynamic or novel situations that demand fine-grained motor control and continuous adaptation and constrains the diversity and compositionality of learned skills.

Second, \modelname{} still relies heavily on supervised datasets and semi-scripted environments for training reasoning and control policies.
Such dependence constrains the model’s generalization to open-ended, long-horizon tasks and truly unstructured environments.

A third limitation lies in test-time latency.
Running an 8B vision-language backbone on a remote server results in roughly $10$ seconds of model inference and about $15$ seconds of end-to-end delay per control step, which restricts deployment to relatively slow and quasi-static scenarios.
This level of latency is inadequate for fast locomotion or highly dynamic human–robot interaction and calls for more efficient model and system designs.

\noindent\textbf{Future work.}
Future work could expand the action space and improve the granularity of action decoding, enabling the model to execute more diverse and context-dependent skills.
Beyond supervised learning, self-supervised and lifelong learning strategies would allow the agent to autonomously acquire new abilities and adapt to evolving environments without extensive human annotation.
On the systems side, we believe the latency gap can be addressed by combining \modelname{} with model compression techniques (e.g., distillation and quantization), smaller backbones, and more aggressive hierarchical control schemes, where high-level CoT planning runs at a low frequency while compact low-level policies execute on-board.
In addition, integrating richer multi-modal sensing, enhancing cross-domain transfer through large-scale pretraining, and extending the framework to dexterous manipulation and collaborative tasks represent promising directions toward a more adaptive, intelligent, and general-purpose embodied agent capable of reliable operation in complex, unstructured real-world settings.